\documentclass[twoside,11pt]{article}

\usepackage{jmlr2e}
\usepackage[utf8]{inputenc}

\usepackage{times}  
\usepackage{helvet} 
\usepackage{courier}  
\usepackage{graphicx} 
\usepackage{natbib}  
\usepackage{caption} 

\usepackage{amsmath,amssymb}
\usepackage{algorithm2e,setspace}
\DeclareMathOperator*{\argminB}{argmin}
\usepackage{subfig}
\usepackage{cleveref}
\usepackage[version=4]{mhchem}
\usepackage{siunitx}

\begin{document}

\title{Ask-n-Learn: Active Learning via Reliable Gradient Representations for Image Classification}

\author{\name Bindya Venkatesh \email bvenka15@asu.edu \\
	\addr Arizona State University\\
	Tempe, AZ, USA
	\AND
	\name Jayaraman J. Thiagarajan \email jjayaram@llnl.gov \\
	\addr Lawrence Livermore National Labs\\
	Livermore, CA,USA}

%
\maketitle

\begin{abstract}

  Deep predictive models rely on human supervision in the form of labeled training data. Obtaining large amounts of annotated training data can be expensive and time consuming, and this becomes a critical bottleneck while building such models in practice. In such scenarios, active learning (AL) strategies are used to achieve faster convergence in terms of labeling efforts. Existing active learning employ a variety of heuristics based on uncertainty and diversity to select query samples. Despite their wide-spread use,  in practice, their performance is limited by a  number of factors including non-calibrated uncertainties, insufficient trade-off between data exploration and exploitation, presence of confirmation bias etc. In order to address these challenges, we propose Ask-n-Learn, an active learning approach based on gradient embeddings obtained using the pesudo-labels estimated in each iteration of the algorithm.  More importantly, we advocate the use of prediction calibration to obtain reliable gradient embeddings, and propose a data augmentation strategy to alleviate the effects of confirmation bias during pseudo-labeling. Through empirical studies on benchmark image classification tasks (CIFAR-10, SVHN, Fashion-MNIST, MNIST), we demonstrate significant improvements over state-of-the-art baselines, including the recently proposed BADGE algorithm.
\end{abstract}

\section{Introduction}


The superior performance of data-driven methods, including deep learning, comes at the price of requiring large amounts of labeled data. This can be a critical bottleneck in applications involving time-consuming data acquisition or high labeling costs. Furthermore, fully supervised methods assume access to samples representing the entire data distribution beforehand, thus making it challenging to handle changes in data distribution over time or adapt the learned model when diverse samples are incrementally included into the training process. This has motivated the use of \textit{active} learning techniques that involve humans in the training loop to build predictive models that are more data-efficient.

Broadly, the goal of active learning is to select the most useful samples for expert annotation, from an unlabeled dataset, while adhering to a given labeling budget~\cite{settles2009active}. More specifically, we are interested in settings where the labeling is carried out in batches~\cite{settles2009active}. The data selection process is at the core of active learning methods and can be designed based on a variety of heuristics including prediction uncertainty~\cite{gal2017deep}, diversity~\cite{brinker2003incorporating,sener2017active} and model generalization~\cite{freytag2014selecting,cai2013maximizing}. For a classifier, it is well known that samples near the decision boundary tend to be associated with large uncertainties and hence including them into the training set can help improve the model's generalization~\cite{beluch2018power,settles2009active}. On the other hand, to avoid sampling biases in the learned models, one needs to ensure that the training dataset sufficiently captures the diversity of the inherent data distribution~\cite{sener2017active,geifman2017deep}. Balancing between these two complementary objectives is often referred to as the \textit{exploitation-exploration} trade-off~\cite{bondu2010exploration}. Interestingly, hybrid methods that leverage this trade-off tend to outperform approaches that rely only on either criterion~\cite{hsu2015active,baram2004online}.

While a large class of heuristics exist in the literature for measuring uncertainties and sample diversity, recently, Ash \textit{et al.}~\cite{ash2019deep} showed that gradient embedding of unlabeled data, obtained using pseudo-labels from the model's current state, is highly effective for both capturing uncertainties and measuring diversity. Pseudo-labeling refers to the process of generating \textit{labels} for unlabeled data to drive the sample selection process. Despite their effectiveness in building predictive models under minimal supervision~\cite{ash2019deep}, methods based on pseudo-labeling are known to suffer from the undesirable behavior of confirmation bias when incorrect pseudo-labels are used in the learning process. Note that, this problem arises even in semi-supervised learning and is typically addressed by including additional regularization to the pseudo-labeling process~\cite{arazo2019pseudo}. On the other hand, existing methods that directly utilize uncertainty scores for sample selection, e.g. entropy, are known to perform poorly when the uncertainties are not well calibrated.

In this paper, we propose a new active learning approach, \textit{Ask-n-Learn}, which addresses the inherent limitations with both pseudo-labeling and uncertainty based methods. While our approach uses gradient embeddings, similar to BADGE~\cite{ash2019deep}, for selecting samples, we propose to utilize calibrated uncertainties to produce reliable gradient embeddings, and employ a data augmentation strategy for avoiding confirmation bias during pseudo-labeling. Using benchmark image classification tasks (SVHN, MNIST, Fashion-MNIST and CIFAR-10), and different deep architectures, we show that the proposed approach significantly outperforms existing active learning approaches, including BADGE~\cite{ash2019deep}, both in terms of test accuracy and prediction calibration.

\section{Related Work}


 

The long-standing problem of active learning (AL) is aimed at enabling faster model generalization, in terms of the amount of supervision required.
AL methods can be broadly categorized based on a number of factors including the type of query (e.g., stream based, pool based query etc), sample selection technique, label acquisition strategy etc.~\cite{settles2009active}. More specifically, in terms of the sample selection, there exists two main classes of AL methods, namely diversity  and uncertainty based methods. In the former approach, by promoting diversity, one expects to obtain a minimal subset of samples that covers the variations expected in the entire data distribution. For example,
~\cite{brinker2003incorporating,guo2008discriminative} proposed diversity based heuristics to select a batch of unlabeled samples in AL tasks. In~\cite{wang2015querying}, the authors develop a batch mode AL strategy based on empirical risk minimization principle. 
The AL problem was posed as a core-set selection task in ~\cite{sener2017active}, wherein a new heuristic to select diverse samples, specifically for convolutional networks, was designed. Finally, a more recent class of diverse sampling methods has been based on adversarial learning~\cite{sinha2019variational,ducoffe2018adversarial,gissin2019discriminative,zhu2017generative}. 

On the other hand, uncertainty based AL methods aim at selecting data samples that the model is most likely uncertain about. Several uncertainty heuristics including entropy, confidence and distance-measures have been proposed in the literature for both Bayesian~\cite{bondu2010exploration,gal2017deep} and non-Bayesian settings~\cite{yang2016active}. In~\cite{kirsch2019batchbald}, the authors introduced an acquisition strategy based on  mutual information for Bayesian active learning. Similarly, Bayesian model uncertainties were utilized for AL in high-dimensional data spaces in~\cite{gal2017deep}. AL methods based on uncertainties from ensembles techniques have also been found to be effective~\cite{melville2004diverse, beluch2018power}. Given the challenges in quantifying model and data uncertainties, recently,~\cite{kiyasseh2020alps} investigated the use of data perturbations to design an acquisition function. A task-agnostic loss prediction framework was proposed in~\cite{yoo2019learning} to select samples that the model is most likely to misclassify. Similarly,~\cite{konyushkova2017learning} proposed to train a regressor to predict the expected error reduction for samples, in AL tasks.

In addition to the uncertainty and diversity based methods, hybrid methods combine these heuristics have also been proposed~\cite{yang2017suggestive}. For instance,~\cite{li2013adaptive} proposed an adaptive AL approach to combine the information density and uncertainty measure fors image classification. Sinha \textit{et al.}~\cite{sinha2019variational} proposed a hybrid method by combining variational autoencoders and adversarial learning frameworks. Several existing works have systematically studied the trade-off between using the uncertainty and representative sampling as exploration-exploitation for AL~\cite{loy2012stream,bondu2010exploration,osugi2005balancing}. Hybrid methods based on quantifying the model generalization have become more popular recently~\cite{freytag2014selecting,cai2013maximizing, huang2016active, bouneffouf2016exponentiated}, including the state-of-the-art BADGE algorithm~\cite{ash2019deep}. When compared to existing methods, \textit{Ask-n-Learn} also falls under the category of hybrid methods and leverages gradient embeddings to perform the sample selection. Through crucial modifications to the AL pipeline, namely uncertainty calibration and data augmentation for confirmation bias control, it achieves significantly better convergence compared to existing methods.


\section{Proposed Approach}
\subsubsection{Notations.} Given the input data sample $\mathbf{x} \subset \mathcal{X} \in \mathbb{R}^{d}$, we consider a ${K}$-way multi-class classification task, i.e., predicting the target label $\mathrm{y} \subset \mathcal{Y} \in \{1,2,\ldots,K\}$. In this paper, our goal is to select the most informative query samples to build a classifier ${f}: \mathcal{X} \mapsto \mathcal{Y}$. We assume that the initial seed dataset of size $S$ is drawn randomly from the train dataset and annotated, i.e., $\mathcal{S}= \{(x_i,y_i)\}_{i=1}^S$. Given the total annotation budget $B$, we develop a batch selection strategy, i.e. pick a subset of $b$ samples from an unlabeled pool, $\mathcal{U} = \{\mathbf{x}_j\}_{j=1}^N$, in every iteration. Finally, we use the set  $\mathcal{L} = \{\mathbf{x}_i, y_i\}_{i=1}^M$ to denote the labeled training data, which grows by including newly annotated samples through the algorithm. Note, in the first iteration, $\mathcal{L} = \mathcal{S}$.

\begin{figure}[!t]
	\centering
	\subfloat{\includegraphics[width=0.48\textwidth]{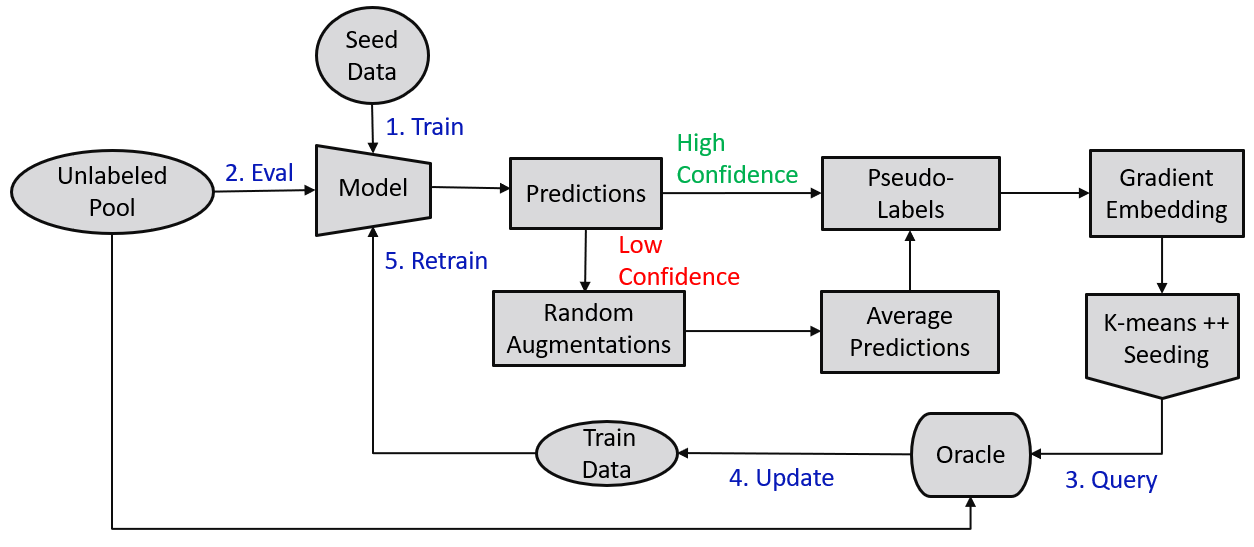}}
	\caption{{An illustration of the proposed active learning approach \textit{Ask-n-Learn}, that utilizes reliable gradient representation obtained via calibrated classifier models and a data-augmentation strategy for reducing confirmation bias.}}
	\label{fig:block_diag}
\end{figure}

\subsection{Formulation}
Similar to the recently proposed BADGE algorithm~\cite{ash2019deep}, our approach utilizes the gradient embeddings to design the sampling heuristic. More specifically, in every iteration, we compute the gradient embeddings for all samples in $\mathcal{U}$ based on the loss function computed with respect to the the pseudo-labels estimated using the current state of the classifier $f$. Note that, pseudo-labels are hard labels obtained by picking the class with the highest softmax probability. In contrast to existing active learning approaches that use explicit predictive uncertainties, the lengths of the gradient vectors provide an estimate of the inherent uncertainties. Finally, similar to~\cite{ash2019deep}, we use the K-means++ seeding algorithm~\cite{arthur2006k} on the gradient embedding to select the query samples, thus promoting the diversity objective.

Despite outperforming existing active learning methods~\cite{ash2019deep}, the success of this sample selection heuristic relies heavily on the quality of the pseudo-labels and the resulting gradient embeddings. Hence, in this work, we advocate for the use of well-calibrated predictive models to obtain reliable gradient embeddings, and we also propose to leverage data augmentation strategies to alleviate the effects of confirmation bias arising due to incorrect pseudo-labeling. As we show in our empirical studies, with these key modifications to the sampling process, \emph{Ask-n-Learn} consistently outperforms BADGE at all sample sizes.

\paragraph{Building Calibrated Classifiers.}Calibration metrics are typically used to measure the consistency between a model's prediction probabilities and the true likelihood (or accuracy). It has been showed in several recent works that by utilizing an explicit calibration objective during training can lead to highly reliable predictive models~\cite{nixon2019measuring,guo2017calibration}. In this paper, we show that such a calibrated model is necessary to obtain effective gradient embeddings for subsequent sampling. Formally, the training objective to build a well-calibrated classifier can be expressed as

\begin{equation}
    \hat{\theta} = \argminB_\theta  \ell_f = \ell_\mathrm{ce}+ \lambda  \ell_\mathrm{calib},
\end{equation}where $\ell_{f}$ represents the overall loss that needs to be minimized using the labeled training data $\mathcal{L}$. This objective is a combination of the standard cross-entropy loss $\ell_\mathrm{ce}$ and a calibration objective $\ell_\mathrm{calib}$ with a regularization weight $\lambda$.


In this work, we consider two different calibration objectives, namely the Variance Weighted Confidence Calibration (VWCC) and the Likelihood Weighted Confidence Calibration (LWCC). Note this regularized optimization eliminates the need for a separate calibration dataset, as required by post-hoc calibration strategies~\cite{guo2017calibration}. 

\noindent \textit{(a) Variance Weighted Confidence Calibration (VWCC)}: In this strategy we employ stochastic inferences to capture the epistemic uncertainties to calibrate the classifier's confidence in its predictions. In particular, we utilize the objective introduced in~\cite{seo2019learning}, where the variance across multiple stochastic inferences are used to regularize the optimization. Here, we utilize a label-smoothing regularization with penalty $\lambda$ specified as the measured variance. Mathematically, this is as follows:

    \begin{align}
    \nonumber {\ell}_{VWCC} &= \frac{1}{M}\sum_{i=1}^M (1 - \alpha_i){\ell}_{ce}^i + {\ell}_{calib}^i \\
    \nonumber &=\frac{1}{M} \sum_{i=1}^M \sum_{t=1}^T -(1 - \alpha_i)\log(p(\hat{\mathrm{y}}^t_i|\mathbf{x}_i)) \\ &\phantom{aaadaaaa}+ \alpha_i D_{KL}({U}(\mathrm{y})||p(\hat{\mathrm{y}}_i^t|\mathbf{x}_i)).
        \label{vwcc}
\end{align}Here, the softmax predictions are obtained via $T$ stochastic inferences for each sample $\mathbf{x}_i$, i.e., $p(\hat{\mathrm{y}}_i^t|\mathbf{x}_i), t=1\cdots T$, and the variance across these predictions $\alpha_i$ is estimated via the Bhattacharyya coefficients. When the variance is high, the loss function is designed to encourage the softmax predictions to be smoothed to an uniform distribution $U$, via using KL-divergence objective. This ensures that the prediction distribution has an higher entropy when the model is more uncertain about its prediction.

\noindent \textit{(b) Likelihood Weighted Confidence Calibration (LWCC)}: We propose a new calibration objective based on the estimated likelihoods, which does not require multiple stochastic inferences. Similar to the VWCC approach, we use the likelihood estimates to control the label smoothing regularization. In particular,

 \begin{align}
    \nonumber &{\ell}_{LWCC} = \sum_{i=1}^M {\ell}_{ce}^i + \lambda [\beta_i D_{KL}({U}(\mathrm{y})||p(\hat{\mathrm{y}}_i|\mathbf{x}_i))], \\
    &\text{where } \beta_i = \bigg(1 - \max(\hat{\mathrm{y}}_i)\bigg)^{\mathbb{I}(\mathrm{y}_i = \hat{\mathrm{y}}_i)}.
        \label{lwcc}
\end{align}When the softmax prediction $p(\hat{\mathrm{y}}_i|\mathbf{x}_i)$ for a sample $\mathbf{x}_i$ is both incorrect and associated with high confidence, those overconfident predictions are adjusted to produce a high-entropy distribution. 

\paragraph{Obtaining Gradient Embeddings.}
Given the unlabeled set $\mathcal{U}$, let $\{{\hat{\mathrm{y}}}_j\}_{j=1}^N$ represent the pseudo-labels predicted by the model $f({\mathcal{U};\theta})$. The gradient embeddings for all $\mathbf{x} \in \mathcal{U}$ are computed as the gradient of the loss function w.r.t the last layer of the network. Note that, even with a calibrated model, we use only $\ell_{ce}$ to compute the gradients. More specifically, the gradient $g_\mathbf{x}$ of a sample $\mathbf{x}$ w.r.t to the last layer $\theta^{o}$ of the network ${f}(\mathbf{x},\theta)$ can be written as

\begin{align}
    \left(g_{\mathbf{x}}\right)_{i}=\left(p_{i}-I(\hat{y}=i)\right) z(\mathbf{x} ; \theta \backslash \theta^{o}),
    \label{eq:grad}
\end{align}where $z$ represents a non-linear transformation applied before the final layer (denote by $\theta \backslash \theta^{o}$). The model's uncertainty about an unlabeled sample is captured via the gradient norm, since the gradient w.r.t to the optimization objective represents the expected model change and its direction.

\paragraph{Reducing Confirmation Bias via Data Augmentation.} We propose to refine the pseudo-label estimates for unlabeled samples, prior to using them in gradient embedding computation. More specifically, given our calibrated predictive model, we retain the pseudo-labels when their confidence (maximum softmax probability) exceeds a preset threshold. In cases when this condition is not satisfied, we obtain more reliable pseudo-label estimates  by averaging predictions over multiple augmented versions of the unlabeled samples. In our implementation, we use pseudo-labels obtained by averaging predictions from $k$ randomly augmented versions of an unlabeled sample (details of the augmentation are given in the experiments section). Note that, existing semi-supervised learning approaches routinely use this idea to produce highly consistent classifier models~\cite{zhu2009introduction,zhu2005semi}. Such a label regularization strategy based on calibrated predictions reduces the confirmation bias, as this does not completely rely on the model's existing belief. 

\paragraph{Exploration-Exploitation Trade-off.}
While designing an active learning heuristic, balancing the trade-off between exploration and exploitation of the data space plays a crucial role in selecting informative samples. In the context of active learning, exploration refers to selecting samples to query from an non-sampled region of data distribution, while exploitation refers to selecting query samples from data space that is already sampled. The task of exploration requires the AL heuristic to select diverse samples that are not already sampled, similarly the exploitation task requires to select samples that the model is most uncertain about and the samples that are difficult to learn from (closer to the decision boundary). As showed in our results, our proposed strategies for model calibration and pseudo-label estimation, combined with the K-means++ seeding achieves significantly better trade-off between these two objectives, thus leading to better quality classifiers for the same sampling budget.

\subsection{Algorithm} The steps involved in the proposed \textit{Ask-n-Learn} method are detailed in Algorithm:~\ref{algo}. Given the labeled seed data, the first step is to build a well-calibrated classifier (equation~\ref{vwcc} or~\ref{lwcc}). At each active learning iteration, the pseudo-labels for $\mathbf{x} \in \mathcal{U}$ are estimated using the proposed strategy. These pseudo-labels are then used to compute the gradient embedding representations, from which a batch of diverse samples are selected via K-means++. Selecting samples with high gradient norms and different directions allows the active learning strategy to consider both the uncertainty and diversity factors while sampling the query points. Assuming that the expert annotates the selected query samples, the model is retrained using the updated labeled set $\mathcal{L}$. This process is repeated until the annotation budget is exhausted.


\RestyleAlgo{boxruled}
\begin{algorithm}[t]
	\setstretch{1.0}
	\KwIn{Seed data $\mathcal{S}$, Unlabeled set $\mathcal{U}$, Query set $\mathcal{Q}$; \newline Labeling budget $B$, Batch size $b$, Confidence threshold $\tau$, Number of augmentations $k$.}
	\KwOut{Trained classifier $f(\theta;\mathcal{L})$}
	\textbf{Initialize}: Parameters $\theta$ of model $f$, Set $\mathcal{L} = \mathcal{S}$\;

	\While{budget $B$ $\neq$ $0$}{
		\For{$\mathbf{x}_i$ in $\mathcal{U}$}{
		    \textbf{Train :} Minimize $ {\mathop{\mathbb{E}}}_{\mathbf{x}\in \mathcal{L}}~[\ell_f(f(\mathbf{x};\theta),y) ]$ using VWCC or LWCC regularization;
		    
			compute $\mathbf{y}_i = f(\mathbf{x}_i;\theta)$ ;\\
		    \eIf{${\rm max} ~p(\mathrm{y}_i|\mathbf{x}_i;\theta)     \ge \tau$}{
                $\hat{\mathrm{y}}_i = {\rm argmax}~p(\mathrm{y}_i|\mathbf{x}_i;\theta)$}
             {
                Obtain $\mathbf{x}_i^{(j)} = \text{Aug}(\mathbf{x}_i), j = 1,\cdots, k$;
                $\hat{\mathrm{y}}_i = {\rm argmax}~\big( \frac{1}{k} \sum_{j=1}^{k} p(\mathrm{y}_i|\mathbf{x}_i^{(j)};\theta) \big)$
             }
		compute gradient embeddings using eqn.~\eqref{eq:grad} and pseudo-label $\hat{\mathrm{y}}_i$;  
		}
    \textbf{Sampling :}	Use K-means++ on the gradient embeddings and select a batch of $b$ samples from $\mathcal{U}$ to form a Query set $\mathcal{Q}$;
    
	\textbf{Update : } $\mathcal{U} \longleftarrow \mathcal{U} - \mathcal{Q}, ~~~~ \mathcal{L} \longleftarrow \mathcal{L} \cup \mathcal{Q}, ~~~~ B \longleftarrow B - b$
	}
	Return the final trained model 
\caption{Ask-n-Learn}\label{algo}
\end{algorithm}

\section{Empirical Studies}

\begin{figure*}[t]
	\centering
	\includegraphics[width=0.99\linewidth,keepaspectratio]{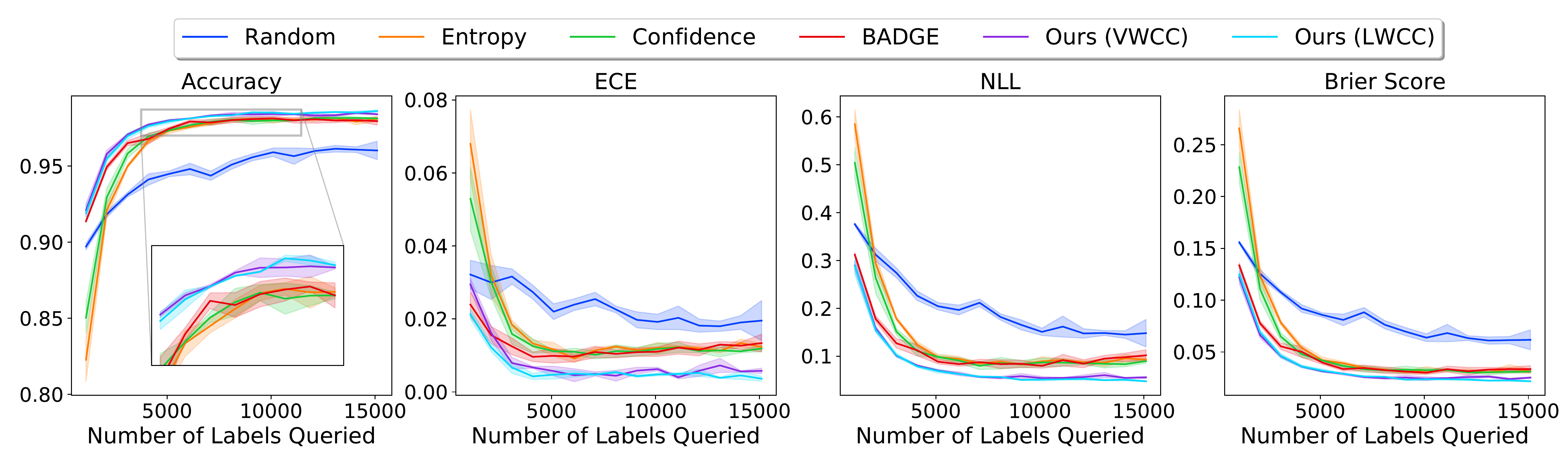}
	\caption{\textit{MNIST}: Generalization performance and reliability analysis of \textit{Ask-n-Learn}. We plot the mean and standard deviation for each of the metrics as training data is added in batches ($b = 1000$).}
	\label{fig:mnist}
\end{figure*}

\begin{figure*}[t]
	\centering
	\includegraphics[width=0.99\linewidth,keepaspectratio]{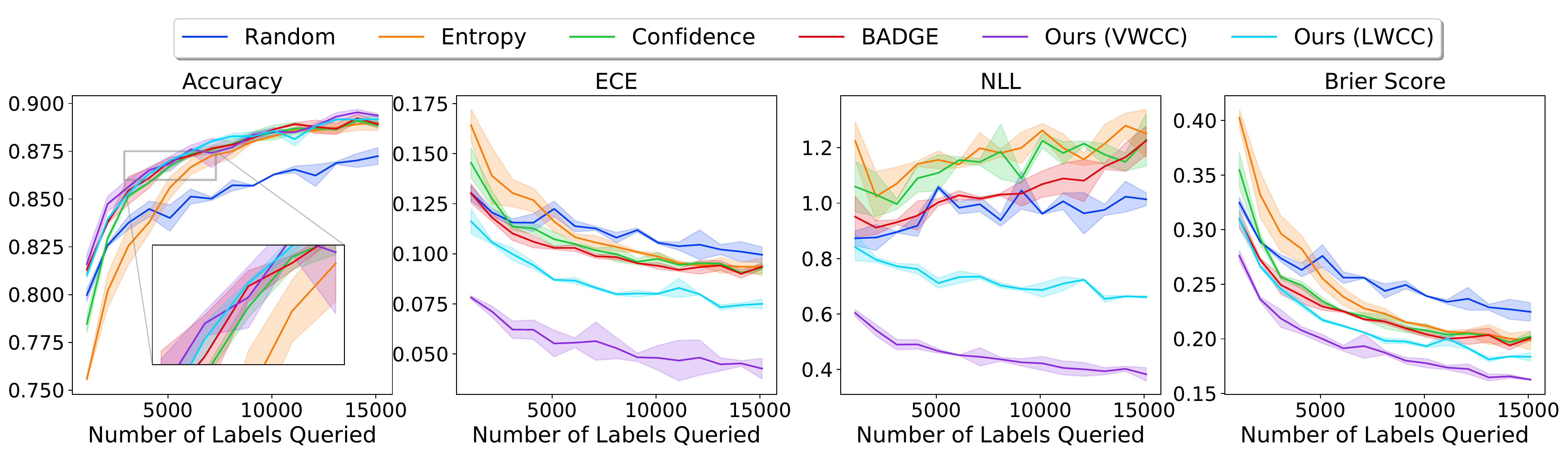}
	\caption{\textit{Fashion-MNIST}: Performance of \textit{Ask-n-Learn} in comparison to the baseline methods. With batch size $b = 1000$, we show the results obtained using a $3-$layer fully connected network.}
	\label{fig:fmnist}
\end{figure*}

\begin{figure*}[t]
	\centering
	\includegraphics[width=0.99\linewidth,keepaspectratio]{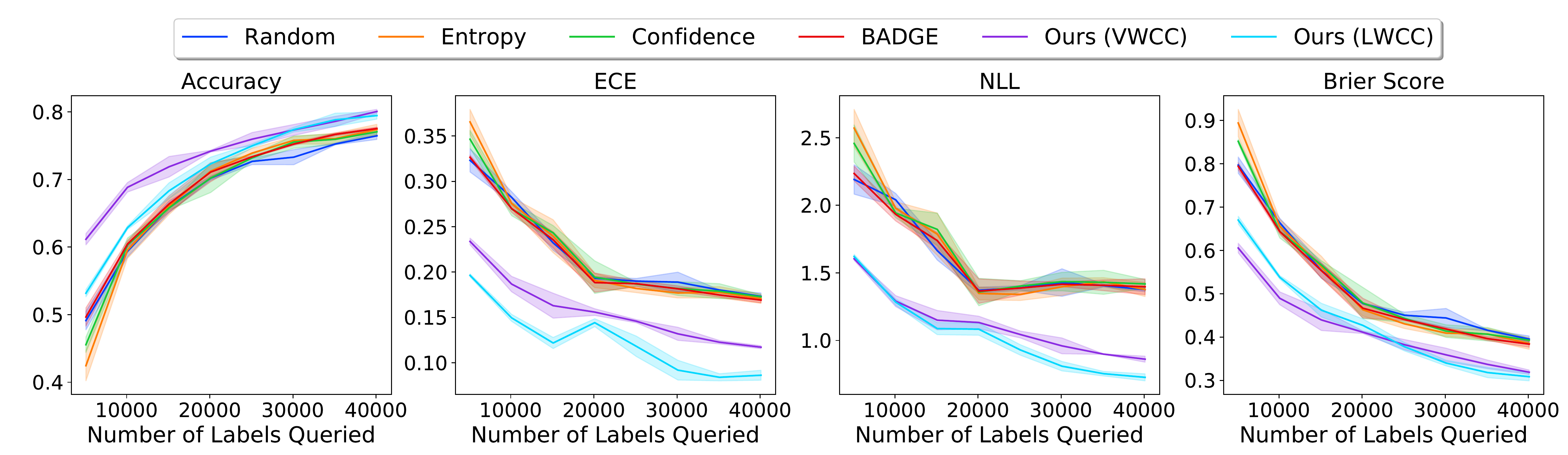}
	\caption{\textit{CIFAR-10}: \textit{Ask-n-Learn} provides significant improvements in terms of both accuracy and calibration metrics, particularly at lower training sizes. In this case, we use the ResNet-18 architecture with $b = 5000$.}
	\label{fig:cifar}
\end{figure*}

\begin{figure*}[t]
	\centering
	\includegraphics[width=0.99\linewidth,keepaspectratio]{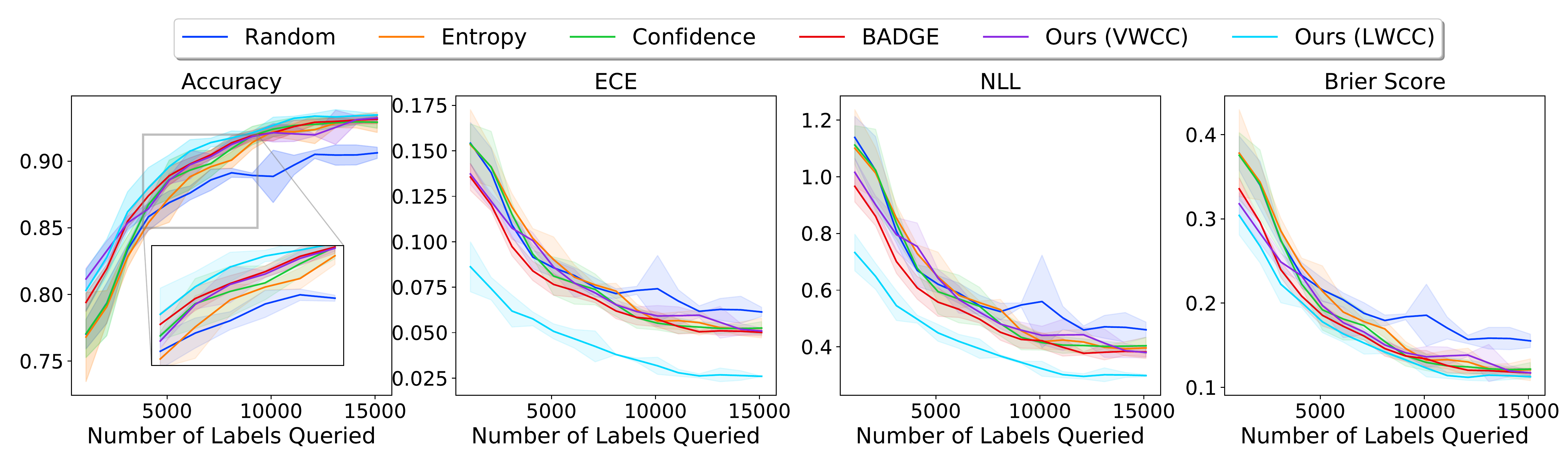}
	\caption{\textit{SVHN}: While even simpler AL heuristics such as entropy and prediction confidence perform comparatively,  \textit{Ask-n-Learn}, in particular LWCC, provides non-trivial performance gains ($1\%-2\%$) when the number o queried labels was low.}
	\label{fig:svhn}
\end{figure*}

\begin{figure*}[!htb]
	\centering
	\subfloat[Noise Ratio = 0.1]{\includegraphics[width=0.95\linewidth,keepaspectratio]{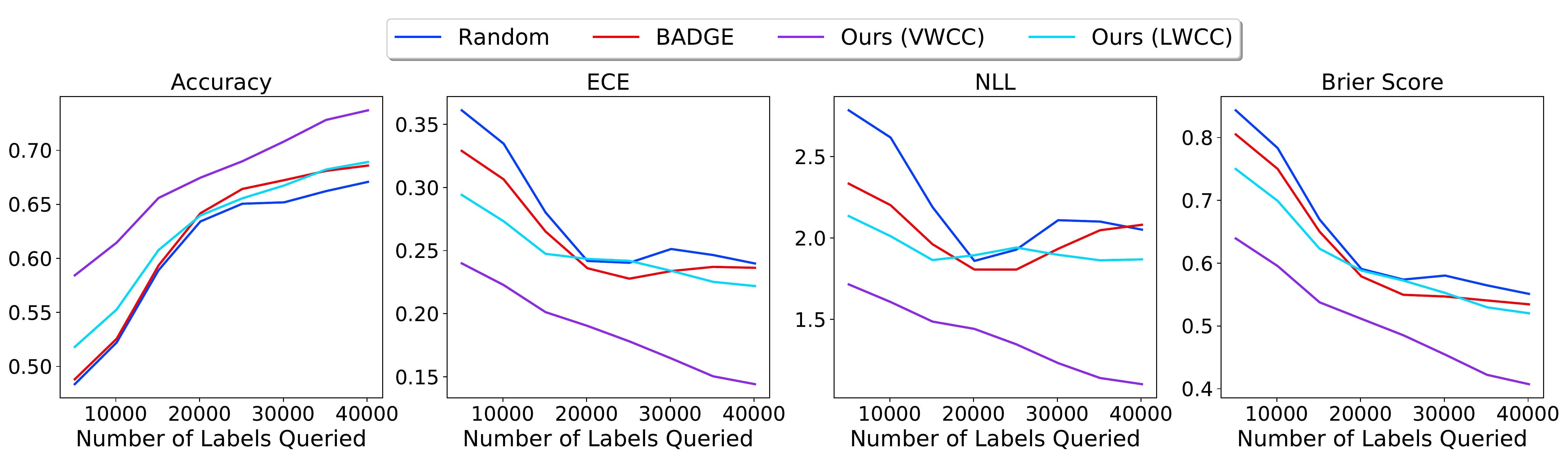} \label{fig:noisy_oracle_01}}
	\vfill
	\subfloat[Noise Ratio = 0.2]{\includegraphics[width=0.95\linewidth,keepaspectratio]{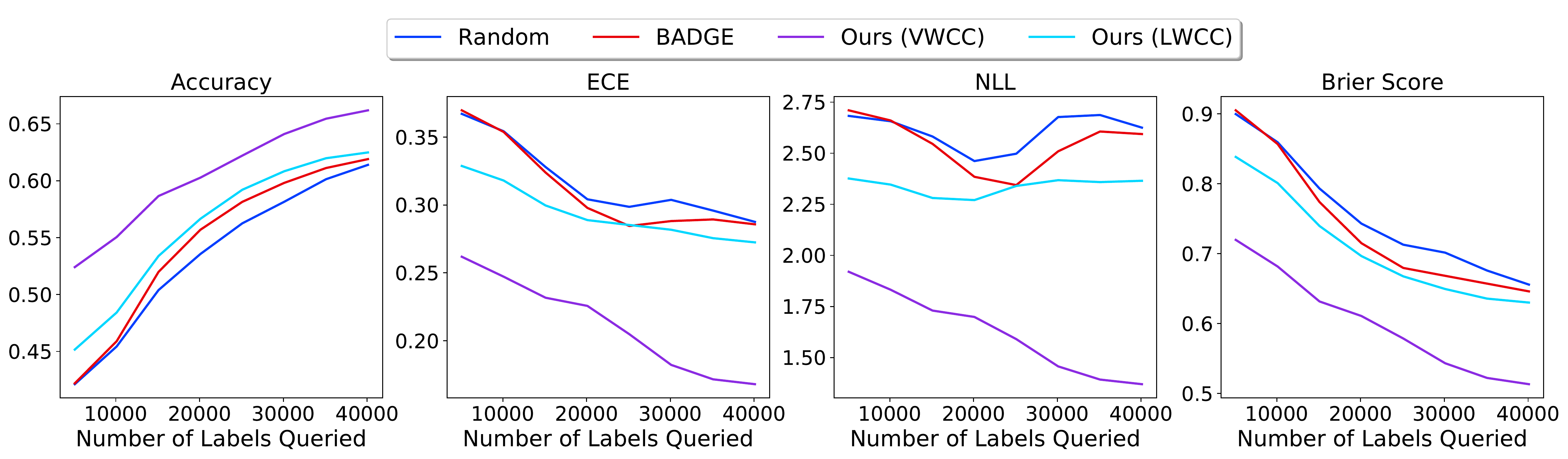} \label{fig:noisy_oracle_02}}
	\caption{Generalization performance and reliability analysis of active learning strategies on CIFAR-10, in the presence of a noisy oracle: (a) $10\%$ and (b) $20\%$ of the correct oracle labels were randomly corrupted in each iteration of the AL algorithm.}
\end{figure*}

In this work, we consider deep image classification tasks and compare the model generalization performance, as a function of the number of labeled training samples utilized, from different active learning methods. In order to evaluate the reliability of the resulting models, we also measure prediction calibration via standard metrics. Finally, we evaluate the robustness of the proposed approach when the oracle is noisy, a commonly encountered challenge in practice. 

\subsection{Experiment setup }
\paragraph*{Datasets and Architectures:} In our study, we consider four benchmark image classification tasks, namely CIFAR-10~\cite{cifar10}, SVHN~\cite{netzer2011reading}, Fashion-MNIST~\cite{xiao2017fashion} and MNIST~\cite{lecun-mnisthandwrittendigit-2010} datasets.
We use the standard ResNet-18~\cite{he2015deep} VGG-16~\cite{simonyan2014very} architectures for CIFAR-10 and SVHN classification experiments respectively. A three layer MLP with configuration $[256-256-256]$ and ReLU activation was used for Fashion-MNIST and MNIST experiments.

\paragraph*{Hyperparameter choices:} We implemented the proposed \textit{Ask-n-Learn} active learning strategy as shown in Algorithm~\ref{algo}. The batch size ($b$) of samples queried in each active learning cycle was fixed to $5$K samples for the CIFAR-10 dataset, and $1$K samples for SVHN, Fashion-MNIST and MNIST. All the models were retrained from scratch in every active learning cycle and trained until convergence. The baselines were implemented following hyperparameter settings in~\cite{ash2019deep} and trained with the standard cross-entropy loss. We used the Adam optimizer with a learning rate of 0.001. All models were trained with an initial random seed data of size $100$. The experiments were repeated for three trials with different random seeds, and average performance along with their variances are reported.

\paragraph*{Baselines:}
The proposed active learning strategy including the two calibration training variations are compared against four baselines including the standard active learning heuristics namely the confidence based methods~\cite{wang2014new,ash2019deep}, entropy based methods. Most importantly, we also compare it against the state-of-the-art BADGE algorithm~\cite{ash2019deep} and na\"ive random sampling. The data selection heuristic in the confidence-based AL method is based on confidences from the model predictions, wherein a batch of $b$ samples with lowest class probability predictions are selected~\cite{wang2014new}. In the entropy-based AL baseline, data samples in an unlabeled pool are ranked according to the entropy of the model softmax predictions. The BADGE algorithm~\cite{ash2019deep} aims to select batches of most informative query samples from an unlabeled pool via sampling based on the directions and lengths of the gradients. The random baseline represents the scenario where a batch of samples are randomly selected from the pool of unlabeled samples for the oracle (or human annotator) to label. Consequently, this represents the conventional fully-supervised training protocol. 

\paragraph*{Evaluation Metrics:} In order to evaluate the model's generalization performance on test data, we use the conventional accuracy metric. The reliability analysis of the trained models are carried out using prediction calibration metrics. In particular, we use the standard metrics namely the expected calibration error (ECE), negative-log-likelihood (NLL) and Brier-scores~\cite{guo2017calibration,ovadia2019can,nixon2019measuring}. The ECE metric meaures the discrepancy between the accuracy and model confidence. This metric is computed by dividing the predictions into equal sized bins and averaging the difference between the accuracy in each bin with its corresponding confidence. 

$$
\mathrm{ECE}=\sum_{b=1}^{B} \frac{N_{b}}{N}\left|\operatorname{acc}\left(\mathcal{B}_{b}\right)-\operatorname{conf}\left(\mathcal{B}_{b}\right)\right|,
$$where $N_b$ represents the number of predictions falling in bin $b$ and $\operatorname{acc}(\mathcal{B}_{b})$ is the accuracy and $\operatorname{conf}(\mathcal{B}_{b})$ the average confidence of the samples in bin $b$. The negative-log likelihood or the cross entropy loss is used to measure the quality of the models probabilistic predictions and hence its uncertainties. Finally, the Brier score is computed as square of the difference between the models' softmax predictions and the true labels, and often used to measure the accuracy of the probabilistic predictions.

\subsection{Results and Findings}
\paragraph{Generalization Performance.}
The efficacy of an active learning heuristic is measured via the generalization performance of the trained model in terms of labeling efforts. Hence, we evaluate the different active learning methods through convergence plots of accuracy versus the number of labeled training data utilized to achieve it. As observed in Figure~\ref{fig:mnist}, our proposed approach with both LWCC and VWCC calibration variations performs better than the baselines on the MNIST classification task. More specifically, the convergence of our approach in terms of accuracy is achieved faster when compared to other active learning heuristics. We observe a similar behaviour in Fashion-MNIST (Figure~\ref{fig:fmnist}), CIFAR-10 (Figure~\ref{fig:cifar}) and SVHN (Figure~\ref{fig:svhn}) classification tasks, where our proposed approach consistently outperforms the baselines including the BADGE. For instance, in the CIFAR-10 experiment, test accuracy of about $70$\% was achieved with just around $10$K samples with our VWCC based method when compared to baselines that require nearly double the sample size to achieve the same accuracy. This clearly evidences the effectiveness of \textit{Ask-n-Learn} in selecting the most informative samples for improved generalization. Additionally, by enabling a principled characterization of prediction probabilities for computing the gradient embeddings, we observe that the stochastic inferencing based calibration strategy (VWCC) appears to be superior to the likelihood based calibration (LWCC), except on the SVHN data where the latter achieves faster convergence. These observations clearly illustrate the importance of using reliable gradients combined with the data augmentation strategies that alleviates the effects of confirmation bias in the training pipeline. 

\paragraph{Reliability Analysis.} While accuracy is a widely-adopted metric for model evaluation, it becomes critical to analyse the reliability of deep predictive models. For instance, a model associated with overconfident probabilities while making an incorrect prediction is undesirable. Hence, in addition to evaluating the accuracy of AL methods, we assess the reliability of the trained models by measuring the consistency between model accuracy and the prediction probabilities (ECE, NLL, Brier score). We observe from~\Cref{fig:mnist,fig:fmnist,fig:cifar,fig:svhn} that the proposed strategy consistently produces well calibrated predictive models across all the datasets. In terms of all three metrics, \textit{Ask-n-Learn} produce significantly lower calibration errors compared to the baselines. 

\paragraph{Robustness to a Noisy Oracle.}Active learning heuristics are designed with an assumption that the training labels are annotated accurately. However, in practice, this assumption may be violated and in those cases, the labeling noise (imperfect oracle) can impact the performance of any active learning pipeline. Hence, it is critical to analyze the performance of AL methods in the presence of a noisy oracle. To this end, we repeat the CIFAR-10 classification experiments with varying levels of labeling noise. To simulate such a scenario, a pre-specified ratio of query samples are wrongly labeled in every iteration of the AL process. In particular, we perform experiments where $10$\% and $20$\% of the true labels are randomly replaced with incorrect ones and used for training. From~\Cref{fig:noisy_oracle_01,fig:noisy_oracle_02} it can be seen that the models designed using the proposed active learning heuristic, in particular VWCC, is significantly more robust to labeling noise. More interestingly, during the initial few iterations, the performance of BADGE is similar to that of na\"ive random sampling. In comparison, \textit{Ask-n-Learn} achieves $>10\%$ higher accuracy than the baselines in both $10\%$ and $20\%$ noise cases. This implies that the effects of confirmation bias and predictive miscalibration are reduced through the proposed strategies, thus leading to improved sample selection. 


\section{Conclusions}

In this work, we proposed an active learning framework \textit{Ask-n-Learn}, based on reliable gradient representations for deep image classification tasks. 
We addressed the inherent limitations of both pseudo-labeling and uncertainty scoring based active learning methods, via prediction calibration and data-augmentation based pseudo-label modification. We demonstrated the sample efficiency of \textit{Ask-n-Learn} across various benchmark classification tasks and model architectures, in terms of both accuracy and empirical calibration metrics. Finally, we studied the robustness our approach in an imperfect oracle scenario, and showed that it generalized better when compared to the popular baselines.

 \section{Acknowledgements}
This work was performed under the auspices of the U.S. Department of Energy by Lawrence Livermore National Laboratory under Contract DE-AC52-07NA27344. 

\begin{quote}

\nocite{*}
\bibliography{references}
\bibliographystyle{new-aiaa}

\end{quote}

\end{document}